\def\year{2019}\relax
\documentclass[letterpaper]{article} 
\usepackage{aaai19}  
\usepackage{times}  
\usepackage{helvet}  
\usepackage{courier}  
\usepackage{url}  

\usepackage[dvipdfmx]{graphicx}
\usepackage{bmpsize}

\usepackage{booktabs}
\usepackage{booktabs}
\usepackage{multicol}
\usepackage{epsfig,epsf,subfigure}
\usepackage{siunitx}
\usepackage{url}
\usepackage{xcolor}
\usepackage{todonotes}
\usepackage{makecell}

\begin{document}
%
\title{Context, Attention and Audio Feature Explorations\\ for Audio Visual Scene-Aware Dialog}


\author{Shachi H Kumar \qquad Eda Okur  \qquad Saurav Sahay \AND \qquad Juan Jose Alvarado Leanos \qquad Jonathan Huang \qquad Lama Nachman\\
\\
Anticipatory Computing Lab\\
Intel Labs\\
\texttt{\{shachi.h.kumar, eda.okur,  saurav.sahay,  juan.jose.alvarado.leanos, jonathan.huang, }\\
\texttt{lama.nachman\}@intel.com}\\
}



\maketitle
\begin{abstract}
With the recent advancements in AI, Intelligent Virtual Assistants (IVA) have become a ubiquitous part of every home. Going forward, we are witnessing a confluence of vision, speech and dialog system technologies that are enabling the IVAs to learn audio-visual groundings of utterances and have conversations with users about the objects, activities and events surrounding them. As a part of the 7th Dialog System Technology Challenges (DSTC7), for Audio Visual Scene-Aware Dialog (AVSD) track, We explore `topics' of the dialog as an important contextual feature into the architecture along with explorations around multimodal Attention. We also incorporate an end-to-end audio classification ConvNet, AclNet, into our models. We present detailed analysis of the experiments and show that some of our model variations outperform the baseline system presented for this task. 
\end{abstract}

\section{Introduction}
Training end-to-end dialog systems have obviated the need for building extensive engineered pipelines for various components of a Spoken Dialog System (SDS) \cite{DBLP:journals/jmlr/CollobertWBKKK11} \cite{serban2016building}. Recent progress in visual grounding techniques \cite{yu2016video} \cite{7410636} is enabling machines to understand semantic concepts that are shared across audio, language and vision modalities. Similar progress have been made in the domain of Audio Understanding \cite{hershey2017cnn} allowing machines to listen to the environment and understand various sensory events in the environment. With audio and visual grounding methods, end-to-end multimodal SDSes are now being trained to meaningfully communicate with us in natural language about the real dynamic audio-visual sensory world around us. Real world visual language interactions with machines are still in early stages of demonstrations and have been made possible due to Deep Learning based advancements in fields such as Machine Reading Comprehension, Neural Machine Translation (NMT), Visual Question Answering (VQA) and Visual Dialog \cite{Das_2017}. 

Topic models allow unsupervised learning of latent distributions of vector spaces for text documents. These models can help the SDS to consider additional information coming from the context of question topics and dialog history topics while decoding the answers. In this work, we incorporate context information of the dialog - question, history of questions and answers, and captions in terms of adding underlying topical information expressed in these. We also build on top of the AVSD baseline model \cite{DBLP:journals/corr/abs-1806-08409} by adding different configurations of attention layer to the dialog history LSTMs and audio/aideo aeatures to learn to attend to them while generating answers. The main difference of this approach from the baseline encoder-decoder is that it does not attempt to encode the dialog history purely as a recurrence layer. Instead, it encodes the history into a sequence of learnt vectors and chooses a subset of these vectors adaptively while decoding the answer. This allows the decoder to adaptively cope better with long interactions and not rely purely on the history sentence LSTM to try to remember long length interactions. We develop and test our approaches on the AVSD dataset \cite{DBLP:journals/corr/abs-1806-00525} released as a part of the DSTC7 task.

\section{Related Work}
The encoder-decoder paradigm \cite{sutskever2014sequence} is the most commonly used paradigm for AVSD like applications and was initially developed for the Neural Machine Translation(NMT) domain. While the encoder maps the source encodings to fixed length vectors, the decoder translates the fixed length vectors to the desired target encodings. Many network variations have been proposed as part of this encoder decoder architecture \cite{DBLP:conf/ssst/ChoMBB14}. Attention-based models \cite{DBLP:journals/corr/BahdanauCB14} can dynamically retrieve relevant pieces of the source via selective reading through a relatively simple matching operation. Recent architectures have proposed use of Transformers \cite{DBLP:conf/nips/VaswaniSPUJGKP17} that extend use of Attention mechanism in various interesting ways. Early Deep Learning based approaches in visual grounding (visual-to-text translation) used RNNs for image captioning \cite{Vinyals_2017}. Early Video Captioning work simply extended the image captioning algorithms by average pooling the video frames \cite{DBLP:conf/naacl/VenugopalanXDRM15}. This simple extension fails to capture multiple events spanning longer duration videos. People developed more sophisticated ways to capture longer duration videos by using recurrent encoders for frames \cite{Venugopalan_2015} or use of Attention \cite{7410869} mechanism to selectively focus on relevant video features. Visual Question Answering (VQA) \cite{7410636} further extended the Video Captioning and Neural Machine Translation work and incorporated question-to-answer generation via end-to-end encoder-decoder framework. Visual Dialog task \cite{Das_2017} further extended VQA for multi-turn conversational dialog about static images addressing dialog handling challenges. AVSD task brings together all of these technologies for holding conversations about videos. To capture objects, events and other temporal information from audio and video streams, a lot of recent progress has been made in audio and video understanding area \cite{Hori_2017}. AVSD system exploits these high level features learnt from various pre-trained networks such as VGGish and C3D \cite{yu2016video}.   

\section{Model Description}
\begin{figure}
\centering
\includegraphics[width=0.5\textwidth]{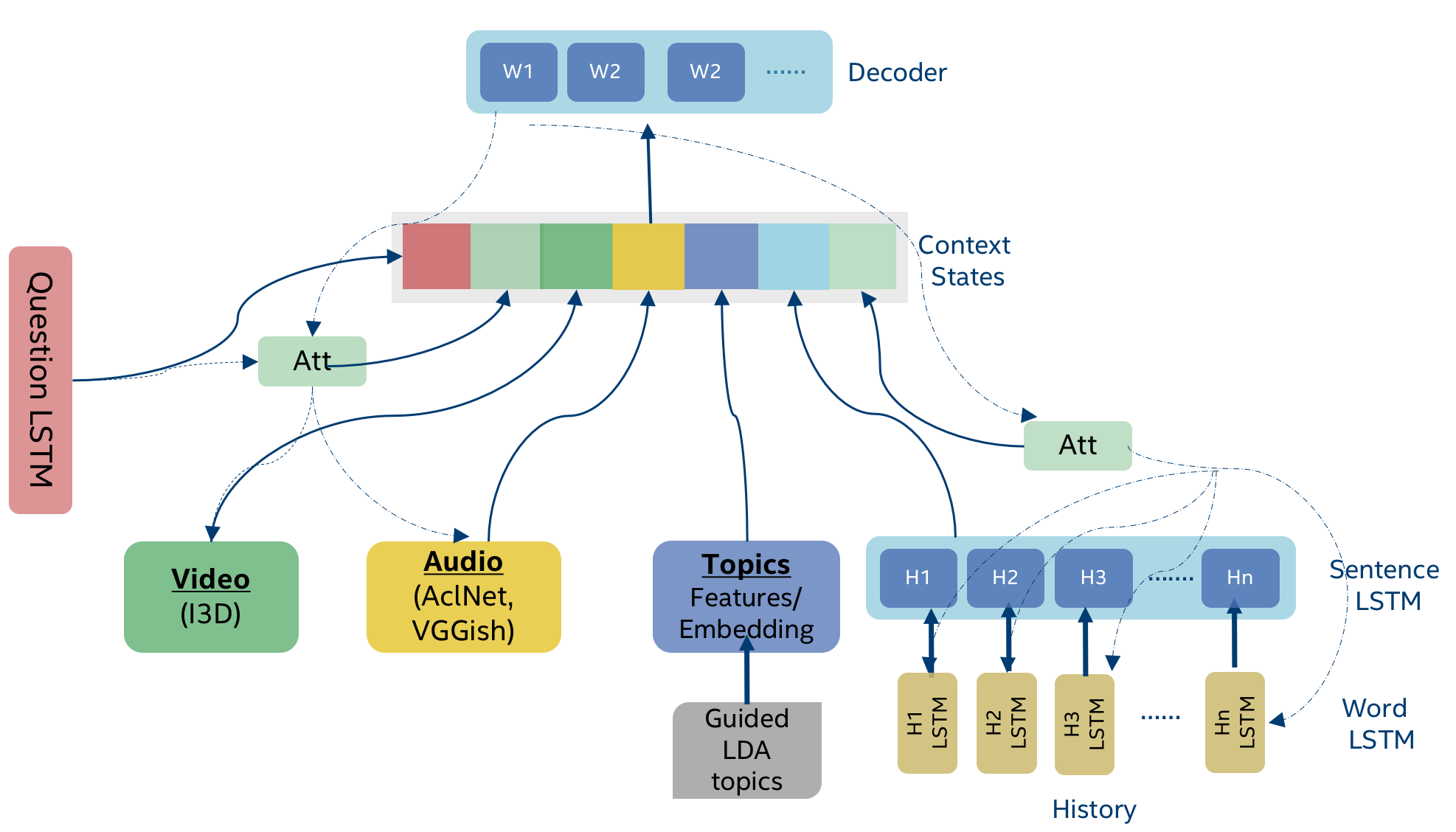}
\caption{Architecture of Our System}
\label{arch}
\end{figure}

In this section, we describe the main architecture explorations of our work. Figure \ref{arch} shows our model with three main architectural extensions as will be discussed in the following sections.

\subsection{Topic Model Explorations}

Topics form a very important source of context in a dialog. A dialog is more likely to contain information about a few specific topics. Charades dataset \cite{DBLP:conf/eccv/SigurdssonVWFLG16} contains videos on common household activities such as watching TV, eating, cleaning, using a laptop, sleeping and so on. Since the dialog data involves the thread of conversation about a given video, there is a natural `topic' of conversation based on the activity in the video. For example, if the video is about a person in the kitchen cooking, the dialog most likely revolves around the things or activities in the kitchen. Similarly, a video about a person watching TV would most likely involve a dialog around the activities in the living room. However, this segregation is not as straight-forward because only one person in the dialog has access to the entire video, and the other person is guessing and questioning about the activity. This demands a computational linguistics approach to understand the topics of the conversation. In this work, we use topic models to map parts of the dialog to topics in an unsupervised way.\\  

\begin{table*}[t]
\begin{center}
  \begin{tabular}{| l | l |}
  \hline
  \scriptsize \textbf{Topic} &\scriptsize  \textbf{Seed Words} \\ 
  \hline
  
  \scriptsize   Entertainment/LivingRoom & \tiny living, room, recreation, garage, basement, entryway, television, tv, phone, laptop, sofa, chair, couch, armchair, seat, picture, sit, play, smile, laugh, watch, listen, turn, ... \\ 
  \hline
  
  \scriptsize   Cooking/Kitchen & \tiny kitchen, pantry, food, water, dish, sink, refrigerator, fridge, stove, microwave, toaster, kettle, oven, stewpot, saucepan, cook, wash, prepare, cut, chop, heat, bake, fry, ... \\
  \hline
  
  \scriptsize   Eating/Dining & \tiny dining, room, table, chair, plate, fork, knife, spoon, bowl, glass, cup, mug, coffee, tea, sandwich, meal, breakfast, lunch, dinner, wine, bar, eat, drink, pour, grab, have, ... \\
  \hline
  
  \scriptsize   Cleaning/Bath & \tiny bathroom, hallway, entryway, stairs, restroom, toilet, towel, broom, vacuum, floor, sink, water, mirror, cabinet, medicine, hairdryer, clean, wash, shower, brush, bath, ... \\
  \hline
  
  \scriptsize   Dressing/Closet & \tiny walk-in, closet, clothes, wardrobe, shoes, shirt, pants, trousers, skirt, jacket, t-shirt, underwear, sweatshirt, coat, hanger, rack, dress, undress, wear, put, fit, hang, try, ... \\
  \hline
  
  \scriptsize   Laundry & \tiny laundry, room, basement, clothes, clothing, cloth, basket, bag, box, towel, shelf, dryer, washer, washing, machine, do, wash, grasp, hold, get, throw, take, pick, dry, ... \\
  \hline
  
  \scriptsize   Rest/Bedroom & \tiny bedroom, room, bed, pillow, blanket, mattress, bedstand, nightstand, commode, dresser, bedside, lamp, nightlight, night, light, lie, snuggle, sleep, rest, tidy, awaken, ... \\
  \hline
  
  \scriptsize   Work/Study & \tiny home, office, den, workroom, room, garage, basement, laptop, desktop, computer, pc, monitor, mouse, keyboard, phone, desk, chair, light, work, study, write, type, ... \\
  \hline
  
  \scriptsize   Sports/Exercise & \tiny recreation, room, garage, basement, hallway, stairs, gym, fitness, floor, bag, towel, ball, treadmill, bike, rope, mat, run, walk, workout, exercise, cycle, lift, jump, jog, ... \\
  \hline
        
  \end{tabular}
    \caption{Sample of Seed Words for 9 Topics}   
    \label{table:seeds9}
\end{center}
\end{table*}

\subsubsection{\textbf{Latent Dirichlet Allocation (LDA).}} Latent Dirichlet Allocation \cite{DBLP:conf/nips/BleiNJ01} is one of the very commonly used topic models to determine the topic distribution of a document. We train topic models on questions, answers, QA pairs, captions and history. The topic distributions generated by these models are incorporated as features into our models or used to learn topic embeddings.  \\

\subsubsection{\textbf{Guided LDA\footnote{https://github.com/vi3k6i5/GuidedLDA}.}}
Since we are interested in identifiying specific activities based on domain knowledge, we use Guided LDA \cite{DBLP:conf/eacl/JagarlamudiDU12} that presents a way to guide topic models to learn topics of specific interest to a user. This technique uses a set of seed words provided by the user, representative of the topics in a corpus.

\begin{itemize}

\item \textbf{\textit{Seed words}}: We derive the seed words for various topics based on the analysis presented in \cite{DBLP:conf/eccv/SigurdssonVWFLG16} for the Charades dataset. We consider topics such as entertainment, cooking, cleaning, resting, etc. A detailed list of seed words for the 9-topics configuration is presented in Table~\ref{table:seeds9}. We obtain the seed words by identifying a set of most common nouns (objects), verbs, scenes, and actions from the Charades dataset analysis.

\item \textbf{\textit{Examples of text with topics}}: Below are some of the examples with the predicted topics (having the highest probability within document-topic distribution vector) for given captions, questions and answers from the AVSD dataset:

\underline{Caption}: \textit{a person smiles as they cook food on the stove . the person takes out their laptop and consults a recipe .}\\
$\rightarrow$ Predicted Topic: \textit{Cooking / Kitchen}

\underline{Question}: \textit{does he take something out of the fridge ?}\\
\underline{Answer}: \textit{yes he takes something out of the refrigerator}\\
$\rightarrow$ Predicted Topic: \textit{Cooking / Kitchen}

\underline{Caption}: \textit{a person runs into the bedroom , undressing . the person checks their phone , turns off the light , then jumps in bed .}\\
$\rightarrow$ Predicted Topic: \textit{Rest / Bedroom}

\underline{Caption}: \textit{a person is playing with their camera while sitting on the stairs . the person goes to take a picture of themselves , but sneezes just as the flash goes off .}\\
$\rightarrow$ Predicted Topic: \textit{Entertainment / Living Room}

\underline{Caption}: \textit{a person walks into the bathroom and turns on the light . the person drinks from a cup of coffee . the person watches themselves in the mirror and smiles .}\\
$\rightarrow$ Predicted Topic: \textit{Cleaning / Bath}

\end{itemize}

\subsection{Attention Explorations}
In the AVSD baseline architecture, the context vector is the main bridge between the multimodal encoder and the answer decoder. The dialog history encoder encodes information via word level LSTMs followed by sentence level LSTMs. The sentence level LSTM may not be able to retain information from earlier sentences in the dialog history. Only the last hidden state of sentence level LSTM is passed on to to the context vector for the decoder to leverage information from dialog history. This single n-dimensional vector carries the burden of retaining useful information from the entire dialog history. We add attention layer to the dialog history representations and audio/video features to selectively focus on relevant parts of the dialog history and audio/video features at every step of the decoder. We calculate the attention weights corresponding to every dialog history turn, the multimodal features and the decoder representation and apply the weights to the history and multimodal features to compute the relevant representations. These help create a weighted combination of the dialog history and multimodal context that is richer than the unweighted single context vectors of the encoder hidden representations. We append the input encoding alongwith the AV multimodal feature encodings and pass it to the decoder LSTM for learning the output encodings.

\subsection{Audio Feature Explorations}
Following the successes of image classification, ConvNets have become popular for audio classification. \cite{hershey2017cnn} showed that large image classification ConvNets trained with huge amount of weakly labeled Youtube data leads to semantically meaningful representations, which has been effectively used for transfer learning in new tasks.  For the sound recognition pipeline, we used an end-to-end audio classification ConvNet, called AclNet \cite{AclNet}. AclNet takes raw, amplitude-normalized 44.1 kHz audio samples as input, and produces classification output without the need to compute spectral features.  The network consists of two separate building blocks: the low-level features which uses two 1D convolutions that produces spectral-like outputs, followed by the high-level features which is inspired by the VGG architecture. 
 
AclNet is trained using the ESC-50 \cite{piczak2015dataset} corpus, which is a dataset of 50 classes of environmental sounds organized in 5 semantic categories (animals, interior/domestic, exterior/urban, human, natural landscapes).  Each sound file lasts for 5 seconds and there are 2000 audio recordings in total. ESC-50 has been widely adopted by the machine audition community as a testbed for novel architectures. At the time of this writing, AclNet shows the best one shot accuracy of~85.65\% (behind an ensemble model) for the ESC-50. 
 
The nature of training data and the classification architecture have several differences from the VGGish model, which is why we believe it provides a diversifying if not complementary effect for the model fusion.  The files and labels of the ESC-50 are carefully curated, as opposed the weakly labeled Youtube data that VGGish is trained with.  The “spectral features” learned by AclNet was completely data-driven, thus it may behave somewhat differently than the mel-spectral features from VGGish.





\section{Dataset}

We use the official-training, prototype-validation and prototype-test sets provided as a part of the DSTC7 Audio Visual Scene-Aware Dialog challenge. Table~\ref{table:dataset} shows the distribution of AVSD data across these different sets. 

\begin{table}[h]
\small
\begin{center}
    \begin{tabular}{| l | l | l | l |}
    \hline
 & \textbf{Training} &  \textbf{Validation} &  \textbf{Test}  \\ \hline
    \# of Dialogs & 7,659 & 732 & 733 \\ \hline
    \# of Turns & 153,180 & 14,680 & 14,660 \\ \hline
    \# of Words & 1,450,754 & 138,314 & 138,790 \\ \hline
    \end{tabular}
    \caption{Audio Visual Scene-Aware Dialog Dataset}
    \label{table:dataset}
\end{center}
\end{table}

\subsection{AVSD Dataset Analysis}
We analyze the official training set of AVSD dataset to gain a better insight on the dialogs, and we also use that knowledge to create certain subsets of the prototype test set to evaluate our results on these subsets in order to observe the desired effects of our models.  

\begin{itemize}
\item \textbf{Q/A Lengths:} We analyzed the number of words in questions and answers in AVSD, and here are our findings:

\subitem Questions: $\mu =~8.46, \sigma =~3.48, range = [~2,~56]$
\subitem Answers: $\mu =~10.41, \sigma =~5.03, range = [~5,~55]$ 

Unlike similar previous datasets like VisDial \cite{Das_2017} and VQA \cite{7410636}, answers in AVSD are significantly longer and much more detailed. We observed a mean-length of 10.4 words in AVSD compared to 2.9 in Vis-Dial and 1.1 in VQA datasets. This indicates the existence of descriptive and conversational answers in AVSD. Similarly, questions are also lengthy as we see a mean-length of 8.5 words in AVSD compared to 5.1 in Vis-Dial and 6.2 in VQA datasets. 

\item \textbf{Binary vs. Non-Binary Q/A:} In VisDial, binary questions are defined as those starting with `do', `did', `have', `has', `is', `are', `was', `were', `can', `could'. We expanded that keywords set with variations based on the most common 100 first words in AVSD questions, and we found that~61.19\% of the questions are binary in AVSD. Similarly, answers that contain `yes' or `no' are assumed to be binary answers in VisDial, and we expanded the keywords with common variations which strongly indicate that the answer is binary. As a results, we observed that~43.07\% of the answers in AVSD are binary. Among them, only~41.69\% are `yes' within all binary (yes/no) answers. We split the dataset into two for binary and non-binary questions, and we did the same for the answers.

\item \textbf{Coreferences in Dialogs:} We analyzed the presence of coreferences in AVSD dialogs by checking whether they contain pronouns like `he', `she', `his', `her', `it', `their', `they', `this', `that', `those', etc. As the AVSD dialogs have 10 turns, the coreferences are quite common, especially towards the later turns as expected. We observed that~62.08\% of the questions and~73.69\% of the answers contain those pronouns, which indicates that the coreference resolution will be an issue and the dialog history should be taken into account more seriously.

\item \textbf{Audio-related Q/A:} Similar to the binary vs. non-binary Q/A, we constructed a list of keywords which strongly indicate that the conversation involves audio-related questions and answers (such as `audio', `sound', `hear', `noise', `voice', etc.). We found that~12.38\% of the questions and~14.39\% of the answers are audio-related in AVSD, and again we created the subsets accordingly.
\end{itemize}

\section{Experiments and Results}

We use the official-training and prototype-validation sets provided for the task  for our experiments and report the results on the prototype-test set. Note that we cannot use the official-test set since the ground truth is not released yet; and we cannot use the official-validation set since it covers the dialogs in the prototype-test set, on which we are reporting our results in this paper. For all of the experiments, we use Adam optimizer and pick the best model based on the lowest validation perplexity for response generation. 

In the subsections below, we present the results of our explorations on topics, attention and audio-based experiments. In general, the CIDEr metric (Consensus-based Image Description Evaluation) \cite{DBLP:conf/cvpr/VedantamZP15} uses sentence similarity and inherently captures the notions of grammaticality, saliency, importance and accuracy. It is also shown to present higher agreement with the judgement of consensus assessed by humans compared to other metrics such as BLEU and ROUGE. Based on this, we focus more on the improvements in CIDEr scores for our experiments.

\subsection{Topic Experiments}
To include topics, we try various configurations. We use separate topic models trained on questions, answers, QA pairs (i.e., concatenated Q and A), captions, history (i.e., concatenated QA pairs until the current dialog turn), and history+captions (i.e., concatenated caption and QA pairs until the current turn) respectively to generate topics for samples from each category. 

\begin{itemize}
\item \textbf{\textit{Standard LDA vs. Guided LDA}}: We experimented with training topic models using both Standard LDA (without seed words) and Guided LDA (with seed words) methods separately for questions, answers, QA pairs, captions, and overall history. 
\item \textbf{\textit{Number of Topics}}: To investigate the effects of variation in number of guided topics on the performance of our models, we experimented with training topic models using 5, 7, 9 and 11 topics. We observed that using 9 seeded topics yields the best scores for the responses. Thus, we continued with 9 topics for all of our topic experiments.
\end{itemize}
In addition, we experimented with adding topics in various ways as described below:

\begin{itemize}
    \item[1] \textbf{Topics as Features}:  We explore incorporating the topic distribution vectors generated by Guided LDA as features for the questions and the dialog history. The question topics are added to the decoder state directly. In one variation, the dialog history topics are copied to all the decoder states directly. For this variation, we explored two configurations to construct dialog history topics: 1) We concatenate QA pair topics and caption topics at each turn, which we directly call our `GuidedLDA' model; 2) We leverage all available topics (i.e., questions, answers, QA pairs, captions, history, history+captions), which we call `GuidedLDA-all'. In another variation, the dialog history topics are added as features to the history encoder LSTM to generate a richer representation, which we call HLSTM-with-topics. We also experiment adding GloVe vector representation \cite{pennington2014glove} for question and history. Specifically, we tried 200 dimension vectors with fine-tuning.  
    \item[2] \textbf{Learning Topic Embeddings}: We learn topic embeddings from topics generated for the questions, QA pairs and captions. 
    For each question, QA pair and caption, we pick the top-3 topics based on the topic proportions generated by the Guided LDA. We try to learn the embeddings for these topics similar to learning word-embeddings. 
\end{itemize}

\begin{table}[h]
\scriptsize
\begin{center}
    \resizebox{\columnwidth+0.05in}{!}{
    \begin{tabular}{| l | l | l | l | l | l | l | l |}
    \hline
      & Bleu1 &  Bleu2 &  Bleu3 &  Bleu4 & Meteor & Rouge & CIDEr \\
    \hline
    Baseline & 
    0.273 & 0.173 & 0.118 & 0.084 & 0.117 & 0.291 & 0.766 \\
    StandardLDA & 
    0.255 & 0.164 & 0.113 & 0.082 & 0.114 & 0.285 & \textbf{0.772} \\
    GuidedLDA & 
    0.265 & 0.170 & 0.117 & 0.084 & \textbf{0.118} & \textbf{0.293} & \textbf{0.812} \\
    GuidedLDA-all & 
    0.272 & 0.173 & 0.118 & \textbf{0.085} & \textbf{0.119} & \textbf{0.293} & \textbf{0.793} \\
    GuidedLDA+GloVe & 
    \textbf{0.275} & \textbf{0.175} & \textbf{0.119} & \textbf{0.085} & \textbf{0.121} & \textbf{0.293} & \textbf{0.797} \\
    Topic-embeddings & 
    0.257 & 0.165 & 0.114 & 0.083 & 0.115 & 0.287 & \textbf{0.772} \\
    HLSTM-with-topics & 
    0.260 & 0.166 & 0.115 & 0.084 & 0.117 & 0.290 & \textbf{0.797} \\
    \hline
    \end{tabular}
    }
    \caption{Topic Experiments}
    \label{table:all-topics}
\end{center}
\end{table}

Table~\ref{table:all-topics} compares the baseline model with the addition of StandardLDA and GuidedLDA topic distributions as features. GuidedLDA performs better than the baseline for the CIDEr, ROUGE and METEOR metrics, while the baseline performs better for the BLEU scores (except BLEU4, which is the same for both). StandardLDA shows better CIDEr score compared to the baseline, but GuidedLDA performs better than StandardLDA for all metrics as expected. Note that when we incorporate all available topics at each turn using GuidedLDA-all model instead of GuidedLDA, we obtain much better BLEU scores; and we reach better or similar performances for all metrics compared to the baseline. GuidedLDA+GloVe is our best performing model and outperforms baseline in all metrics. It also outperforms all other topic-based models that we explored (except GuidedLDA in terms of CIDEr score).  
The HLSTM-with-topics model performs better and similar to the baseline in terms of CIDEr metric and BLEU4, METEOR and ROUGE metrics, respectively. The baseline however performs better than the Topic-embeddings model in all metrics except CIDEr. 

We also compare the GuidedLDA models with the baseline by evaluating on subsets of the dataset consisting of binary and non-binary answers. From Table~\ref{table:topicBinNonBin}, for binary subset, GuidedLDA performs better than baseline for some metrics while GuidedLDA+GloVe shows similar/better performance as compared to baseline for all metrics. Interestingly, for the non-binary subset, our GuidedLDA+GloVe model outperforms baseline for all metrics. This shows that our model can generate better responses for the more complex, non-binary answer types.

\begin{table}
\begin{center}
    \resizebox{\columnwidth}{!}{
    \begin{tabular}{| l | l | l | l | l | l | l | l |}
    \hline
 &\scriptsize Bleu1 & \scriptsize Bleu2 & \scriptsize Bleu3 &\scriptsize Bleu4 &\scriptsize Meteor &\scriptsize  Rouge &\scriptsize  CIDEr \\ \hline
 
 \scriptsize \textbf{Binary} & & & & & & & \\
 \scriptsize  Baseline
& \scriptsize \textbf{0.329}
& \scriptsize {0.220 }
&\scriptsize  {0.156}
&\scriptsize  {0.116 }
&\scriptsize  0.142
&\scriptsize  0.345
&\scriptsize  0.965
 \\

 \scriptsize  GuidedLDA
& \scriptsize 0.318
& \scriptsize 0.216 
&\scriptsize  0.155
&\scriptsize  0.116 
&\scriptsize  \textbf{0.144}
&\scriptsize  \textbf{0.346}
&\scriptsize  \textbf{1.008}
 \\ 
 
  \scriptsize  GuidedLDA+GloVe
& \scriptsize 0.328
& \scriptsize 0.220
&\scriptsize  0.156
&\scriptsize  0.115
&\scriptsize  \textbf{0.146}
&\scriptsize  \textbf{0.347}
&\scriptsize  \textbf{0.995}
 \\ 
 \hline  

  \scriptsize \textbf{Non-binary} & & & & & & & \\
 \scriptsize  Baseline
& \scriptsize {0.233 }
& \scriptsize {0.139} 
&\scriptsize  {0.090}
&\scriptsize  0.060 
&\scriptsize  0.099
&\scriptsize  0.250
&\scriptsize  0.541
 \\ 
 
 \scriptsize  GuidedLDA
& \scriptsize 0.226 
& \scriptsize 0.136 
&\scriptsize  0.089
&\scriptsize  \textbf{0.061} 
&\scriptsize  \textbf{0.100}
&\scriptsize  \textbf{0.253}
&\scriptsize  \textbf{0.590}
 \\ 

 \scriptsize  GuidedLDA+GloVe
& \scriptsize \textbf{0.237} 
& \scriptsize \textbf{0.141} 
&\scriptsize  \textbf{0.092}
&\scriptsize  \textbf{0.063} 
&\scriptsize  \textbf{0.103}
&\scriptsize  \textbf{0.252}
&\scriptsize  \textbf{0.579}
 \\ \hline  

    \end{tabular}
    }
    \caption{Performance on Binary/Non-binary Answers}
    \label{table:topicBinNonBin}
\end{center}
\end{table}

 \begin{table*}[t]
\begin{center}
  \begin{tabular}{| l | l |l | l | l |}
    \hline
\scriptsize \textit{Question:} & \scriptsize \textit{\textbf{is there sound to the video ?}} &\scriptsize 

\textit{\textbf{Did he make a sound when blowing his nose ?}} &   \scriptsize \textit{\textbf{does she say anything ? }}   \\     \hline
\scriptsize Ground Truth &\scriptsize  yes there is sound . nothing important . 
&\scriptsize  he didnt blow his nose &\scriptsize    she does not say anything .   \\     \hline
\scriptsize Baseline + VGGish &\scriptsize no there is no sound in the video  
& \scriptsize  no he did not smile in the video&\scriptsize no there is no sound in the video .\\     \hline
\scriptsize Baseline + AclNet &\scriptsize  yes there is sound in the video 
&\scriptsize  no he didnt sneeze in the video  &\scriptsize  no she does not say anything .  \\     \hline

  \end{tabular}
    \caption{Audio Examples (VGGish vs. AclNet)}    
    \label{table:qualitativeaudio}
\end{center}
\end{table*}

 \begin{table*}[t]
\begin{center}
  \begin{tabular}{| l | l | l | l | l | l | l | l |}
    \hline
 &\scriptsize Bleu1& \scriptsize	Bleu2& \scriptsize	Bleu3& \scriptsize	Bleu4& \scriptsize	Meteor& \scriptsize	Rouge& \scriptsize	CIDEr \\ 
 \hline
 
 \scriptsize Baseline	& \scriptsize \textbf{0.248}	& \scriptsize 0.151& \scriptsize	0.101& \scriptsize	0.071& \scriptsize 0.110 & \scriptsize 0.256 & \scriptsize 0.664\\ 
 
  \scriptsize Word LSTM (all output states)	& \scriptsize 0.223	& \scriptsize 0.138& \scriptsize	0.092& \scriptsize	0.065& \scriptsize
 0.103& \scriptsize	0.262& \scriptsize	0.591  \\ 

 \scriptsize Word LSTM (last hidden states)& \scriptsize	0.229& \scriptsize	0.139& \scriptsize	0.093& \scriptsize	0.065& \scriptsize	0.105& \scriptsize	{0.250}& \scriptsize	{0.661} \\ 

 \scriptsize Sentence LSTM (all output states)& \scriptsize 	0.242& \scriptsize	{0.151}& \scriptsize	\textbf{0.103}& \scriptsize	\textbf{0.073}& \scriptsize	{0.110}& \scriptsize	\textbf{0.261}& \scriptsize	\textbf{0.707 }\\ 
 
  \scriptsize Sentence LSTM (all outputs) +video+audio& \scriptsize 	0.234& \scriptsize	{0.146}& \scriptsize	{0.099}& \scriptsize	{0.070}& \scriptsize	{0.109}& \scriptsize	{0.254}& \scriptsize	\textbf{{0.690 }}\\ 
 
 \hline

  \end{tabular}
    \caption{Decoder Attention over Dialog History and Multimodal Features}    
    \label{table:attn}
\end{center}
\end{table*}

 \begin{table}[t]
\begin{center}
  \resizebox{\columnwidth+0.03in}{!}{
  \begin{tabular}{| l | l | l | l | l | l | l | l |}
    \hline
 &\scriptsize Bleu1& \scriptsize	Bleu2& \scriptsize	Bleu3& \scriptsize	Bleu4& \scriptsize	Meteor& \scriptsize	Rouge& \scriptsize	CIDEr \\ 
 \hline
 \scriptsize   \textbf{Overall} & & & & & & & \\
 \scriptsize   Baseline(B)& \scriptsize	0.273 & \scriptsize	0.173 & \scriptsize	0.118& \scriptsize	0.084& \scriptsize	0.117& \scriptsize	0.291& \scriptsize	0.766 \\ 
\scriptsize    B+VGGish& \scriptsize	0.271& \scriptsize	0.172& \scriptsize	0.118	& \scriptsize0.085& \scriptsize	0.116& \scriptsize	0.292& \scriptsize	\textbf{0.791}\\ 
\scriptsize    B+AclNet& \scriptsize	\textbf{0.274}& \scriptsize	\textbf{0.175}& \scriptsize	\textbf{0.121}& \scriptsize	\textbf{0.087}& \scriptsize	0.117& \scriptsize	\textbf{0.294}& \scriptsize	0.789 \\    
\hline

\scriptsize   \textbf{Audio-related} & & & & & & & \\
\scriptsize Baseline(B)& \scriptsize \textbf{0.267}& \scriptsize 0.179& \scriptsize 0.128& \scriptsize 0.096& \scriptsize 0.120& \scriptsize 0.285& \scriptsize 0.919 \\ 
\scriptsize B+VGGish& \scriptsize 0.266& \scriptsize 0.181& \scriptsize 0.131& \scriptsize 0.099& \scriptsize 0.118& \scriptsize 0.285& \scriptsize 0.907
\\ 
\scriptsize B+AclNet& \scriptsize 0.266& \scriptsize \textbf{0.183}& \scriptsize \textbf{0.132}& \scriptsize \textbf{0.100}& \scriptsize 0.120& \scriptsize \textbf{0.287}& \scriptsize \textbf{0.944} \\ 
\hline
  \end{tabular}
  }
    \caption{Audio Experiments \& Performance on Overall vs. Audio-related Questions}    
    \label{table:audioexp}
\end{center}
\end{table}

\begin{figure}[h]
    \includegraphics[width=.5\textwidth]{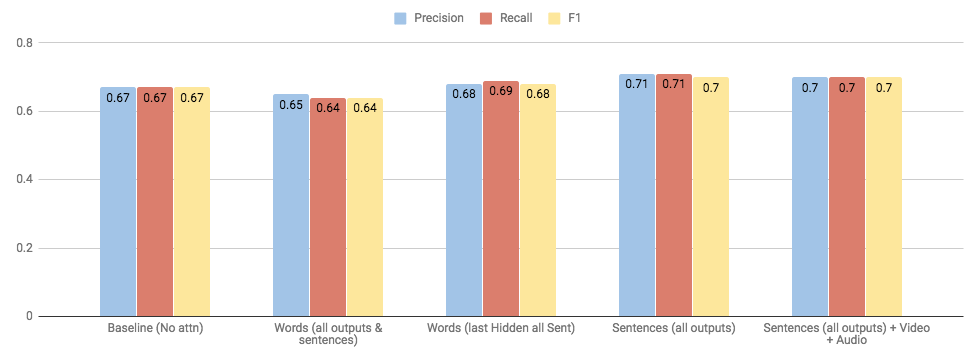}
    \caption{Precision, Recall, F1 Scores for Attention Experiments on Questions Containing Coreferences}
    \label{attentionCoref}
\end{figure}

 

\subsection{Attention Experiments}
In our model, we have modified the baseline architecture and added attention layer with 4 different configurations for the answer decoder to leverage information directly from the dialog history LSTMs and multimodal audio/video features. Since the attention layer pays attention to the dialog history in all our configurations while decoding, we wanted to evaluate the performance of attention with respect to the baseline solely for questions that could benefit from dialog history. It is non-trivial to do such analysis quantitatively. One naive way to accomplish this is to isloate the questions containing coreferences and perform the analysis on these questions. Table \ref{table:attn} shows the performance of our models on the standard evaluation metrics on this coreference-subset of the dataset. 
We further wanted to check how semantically meaningful were the results that we generated using the attention variations. In order to compare the results at a more semantic level, we performed quantitative analysis on dialogs that contained binary answer in the ground truth, by creating the binary answers subset. We evaluate our models on their ability to predict these binary answers correctly (using precision, recall and F1 scores) and present this analysis in Figure \ref{attentionCoref}. 

\begin{itemize}
\item[1] \emph{Attention on Dialog History Word LSTMs, all output states}: In this configuration, we remove the sentence level dialog history LSTM and the decoder computes the attention scores directly between the decoder state and the word level output states for all dialog history. 
We first padded the Word LSTM outputs from Dialog History LSTMs (see Word LSTM in History in Figure \ref{arch}) to the maximum sentence length of all the sentences. We summed up all the attention scores from each of the sentence context vectors with the query decoder state. Directly attending to the output states of the word LSTMs in the dialog history encoder didn't perform well on any of the evaluation metrics compared to the baseline. This attention mechanism possibly attended to way more information than needed. Using this kind of attention, we had hoped that the system could remember answers that were already given (directly or indirectly) in the earlier turns of the dialog. Unfortunately we couldn't perform such quantitative analysis on the dataset in this work. The binary answer evaluation in this setup also performed poorer than the baseline as shown in Figure \ref{attentionCoref}.     
\item[2] \emph{Attention on Dialog History Word LSTMs, last hidden states}: This configuration is similar to the previous configuration with the difference that we only use the last hidden state output representations of the word LSTMs corresponding to the different turns in the dialog. Simpler than the previous setup, we stack up the hidden states from the history sentences for attention computation. From table \ref{table:attn}, this model performs better that the previous setup, slightly worse than the baseline on standard evaluation metrics and slightly better than the baseline on binary answer evaluation from figure \ref{attentionCoref}.
\item[3] \emph{Attention on Sentence LSTM output states}: The baseline architecture only leverages the last hidden state information from the sentence LSTM in the dialog history encoder. Instead, we extract the output states from all timesteps of the LSTM corresponding to $n$ turns of the dialog history. This variation helps the decoder consider all the dialog turn compressed sentence representations via the attention mechanism. As shown in Table \ref{table:attn}, this model performed better than the baseline model on BLEU3, BLEU4, METEOR, CIDEr and ROUGE. Figure \ref{attentionCoref} shows that this model also performed better than the baseline model on binary answer evaluation.
\item[4] \emph{Attention on Sentence LSTM output states and Multimodal Audio/Video Features}: This configuration is similar to the last one with the difference that we add multimodal audio/video features as additional state to the attention module. This mechanism allows the decoder to selectively focus on the multimodal features alongwith the dialog history sentences. This configuration didn't really help improve the evaluation metrics compared to the baseline (see Table \ref{table:attn}) except the CIDEr metric score. The results on binary answer evaluation (Figure \ref{attentionCoref}) improved compared to the baseline but slightly degraded compared to the last configuration.   
\end{itemize}

\subsection{Audio Experiments}
Features extracted using a new state-of-the-art model, Audio Set VGGish were provided as a part of the DSTC7 AVSD challenge. We explore the use of the AclNet features based on the softmax output of the 50-classes from the model described in the previous section.
Table~\ref{table:audioexp} shows the comparison of the baseline model without audio features, and baseline with the addition of the VGGish  and the AclNet features. From the table, Baseline+AclNet (B+AclNet) shows improved scores on all metrics as compared to the baseline and Baseline+VGGish (although B+VGGish gives the best score for this metric). 

We further try to analyse the effects of the audio features specifically on audio-related questions. We split the test dataset into audio-related and non-audio-related dialogs and evaluate the performance of our models on these subsets. From Table ~\ref{table:audioexp}, we observe that B+AclNet performs the best on audio-related questions as well.

Table \ref{table:qualitativeaudio} presents a qualitative analysis of the addition of the VGGish and AclNet features to the baseline model. The question, \textit{``is there sound in the video?"} is incorrectly answered by the B+VGGish model and correctly answered by the AclNet addition. Another interesting observation is made for the question \textit{``Did he make a sound when blowing his nose?"}. B+AclNet refers to ``\textit{sneeze}" in the response which is close to the ground truth answer \textit{``he didnt blow his nose"}, while the VGGish model does not generate a relevant response.

\section{Conclusion and Future Work}
In this paper, we present some explorations and techniques to improve contextual and multimodal end-to-end audio-visual scene aware dialog system. We incorporate context of the dialog in the form of topics, we use various attention mechanisms to enable the decoder to focus on relevant parts of the dialog history and audio/video features, and we incorporate audio features from an end-to-end audio classification architecture, AclNet. As part of the 7th Dialog System Technology Challenges (DSTC7), we validate our approaches on the AVSD dataset and show that some of our models perform better than the baseline on various metrics. Our topic-based contextual models also show that guiding the topic models with seed-words help in improving the overall performance as compared to the baseline.  We also present a quantitative evaluation of our models on their ability to predict binary answers correctly. We show that our attention-based mechanisms perform well on dialog data involving coreference. We also show qualitatively and quantitatively that incorporating our audio pipeline improves the performance as compared to the VGGish features.
AVSD is a new and exiting research area and the current work involved addition of contextual features and Attention mechanisms. As future work, besides fine-tuning the current techniques, we plan to explore multimodal fusion techniques and incorporate better representations for other modalities, such as object, pose and activity recognition algorithms into the pipeline.

\end{document}